\documentclass[12pt]{article}

\usepackage{sbc-template}
\usepackage{graphicx,url}
\usepackage[utf8]{inputenc}  
\usepackage[T1]{fontenc} 
\usepackage{textcomp}
\usepackage{ae} 
\usepackage{amsmath}
\usepackage{amsfonts}
\usepackage{amssymb}
\usepackage{array}
\usepackage{blindtext}
\usepackage{dcolumn}
\usepackage{verbatim}
\usepackage{enumitem}
\usepackage{parskip}
\usepackage[table,xcdraw]{xcolor}
\usepackage{lipsum}
\usepackage{pifont}
\usepackage{afterpage}
\usepackage{imakeidx}
\usepackage{setspace}
\usepackage{pbox}
\usepackage{wrapfig}
\usepackage{titlesec}
\usepackage{dsfont} 
\usepackage{ulem}
\usepackage{bbm}
\usepackage{MnSymbol}
\usepackage{color}
\usepackage{makecell}
\usepackage{footnote}
\usepackage{tablefootnote}

\definecolor{codegreen}{rgb}{0,0.6,0}
\definecolor{codegray}{rgb}{0.5,0.5,0.5}
\definecolor{codepurple}{rgb}{0.58,0,0.82}
\definecolor{backcolour}{rgb}{0.95,0.95,0.92}

\title{Blockchain to Improve Security, Knowledge and Collaboration Inter-Agent Communication over Restrict Domains of the Internet Infrastructure}

\author{Juliao Braga\inst{1,2}, Joao Nuno Silva\inst{2}, Patricia Takako Endo\inst{3,4},\\Jessica Ribas\inst{1}, Nizam Omar\inst{1}}

\address{Universidade Presbiteriana Mackenzie (UPM)
\nextinstitute
IST - INESC ID, University of Lisboa, Portugal
\nextinstitute
Universidade de Pernambuco (UPE), Brazil
\nextinstitute
Dublin City University (DCU), Ireland
\email{\{juliao.braga,joao.n.silva\}@tecnico.ulisboa.pt, patricia.endo@upe.br}
\email{jessica.ribas@mackenzista.com.br, nizam.omar@mackenzie.br}
}

\begin{document} 

\maketitle

\begin{abstract}
  This paper describes the deployment and implementation of a blockchain to improve the security, knowledge and intelligence during the inter-agent communication and collaboration processes in restrict domains of the Internet Infrastructure. It is a work that proposes the application of a blockchain, platform independent, on a particular model of agents, but that can be used in similar proposals, once the results on the specific model were satisfactory.
\end{abstract}

\begin{resumo}
  Este documento descreve o desenvolvimento e implementação de uma \textit{blockchain} para melhorar a segurança, o conhecimento e a inteligência durante os processos de comunicação e colaboração entre agentes em domínios restritos da Infraestrutura da Internet. É um trabalho que propõe a aplicação de uma "blockchain", independente de plataforma, em um mo-delo particular de agentes, mas que pode ser utilizado em propostas similares, uma vez que os resultados no modelo específico foram satisfatórios.
\end{resumo}

\section{Introduction}

\begin{figure}[!ht]
\centering
\includegraphics[scale=0.39]{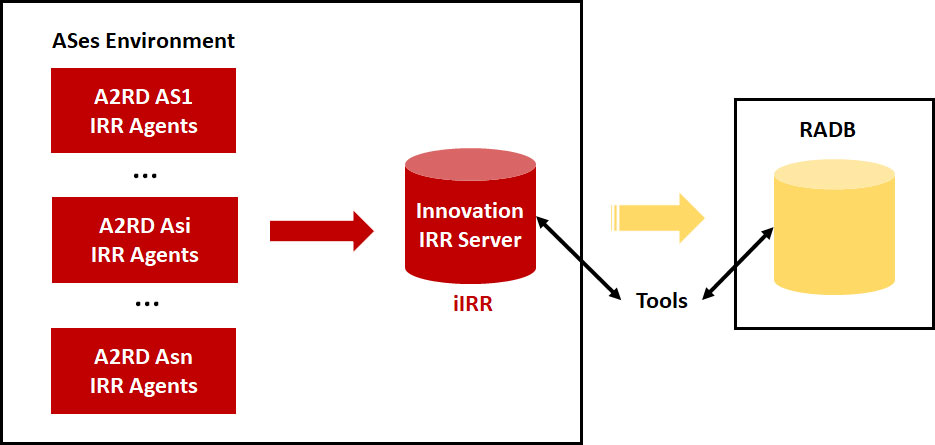}
\caption{The innovation IRR model established by A2RD}
\label{fig:innovationIRR}
\end{figure}

\textit{Autonomous System} (AS) is the name given to the networks making up the Internet \cite{RFC1930:1996}. ASes establish interconnections through a protocol called \textit{Border Gateway Protocol} (BGP) \cite{RFC4271}. BGP is a complex protocol that requires a lot of knowledge from the administrators of an AS. Sometimes the human being also forgets to update information, especially those related to routing policy and that reside on important servers such as \textit{Internet Routing Registry}\footnote{http://www.irr.net/} (IRR), for example. IRR is a distributed database of route and route-related information \cite{Braga2010}. The sometimes neglected participation of the human being during the creation and update IRR objects processes, was the motivation for creating a model of agents which could replace the human interventions (made by email). For this reason, was propose the \textit{Autonomous Architecture Over Restricted Domains} (A2RD) into the restricted domain of an AS, applying as use case over the IRR \cite{Braga:2015}. A2RD replaces the human with your agents, \textit{Intelligent Elements} (IEs), establishing a new IRR model, named \textit{innovation IRR} (iIRR), shown in Figure \ref{fig:innovationIRR}.   

A2RD specialized IEs, automatically create objects as defined by the \textit{Route Policy Specification Language} \cite{RFC2622, RFC4012}. Those objects that can not be created automatically will receive support from AS administrators through a human-computer cooperation mechanism. Nothing is changed in relation to the present and future IRR structure, characterized by the expectations recommended by the stakeholders to the \textit{Internet Engineering Task Force}\footnote{https://ietf.org/} (IETF) and \textit{Internet Research Task Force}\footnote{https://irtf.org/} (IRTF) disseminated through of their formal documents \cite{RFC2650, RFC2725, RFC3707, RFC7682, RFC7909}. Neither does it affect the security concerns surrounding the IRR and Internet governance \cite{Kuerbis:2017}. Similarly, tools that use IRR databases can be used without any modification. A very useful, among others, is the IRR Powertools\footnote{https://github.com/6connect/irrpt}.

For this paper, \textit{blockchain} is a data structure whose components are chained, with guarantee of immutability of its contents, and consequent integrity of the chain preserved by a cryptography process, with difficult computational reversibility. This definition is much simpler but more computationally oriented than those in which blockchain is associated with crypto-economics or crypto-currencies, and often have confusing definitions, but when it is clear, blockchain is defined as a database \cite{Nakamoto2008, Pilkington:2015}.

On the other hand, by abstracting from property of immutability, the data structure like blockchain is a well-known concept used in computer research and originated in the academic literature of the 1980s and 1990s \cite{narayanan2017bitcoin}. As a simple data structure, for example, in works involving \textit{provenance}, which is used as complementary data documentation containing the description of 'how', 'when', 'where', 'why' the data were obtained and 'who' obtained it \cite{braga2008data}. The blockchain model proposed in 2008 to meet the Bitcoin virtual currency has effectively aroused the interest of the research community mainly by the immutability property that ensures data integrity \cite{prusty2017building, bashir2017mastering}. Immutability and integrity are obtained by a hash encryption mechanism \cite{bakhtiari1995cryptographic, rogaway2004cryptographic}. The combination of these two factors and characteristics associated to the blockchain recommended the application in the A2RD model, with the aim of enhancing communication and collaboration among the IEs \cite{Braga:2017}. This proposal is more simpler than those application of blockchain in Internet Infrastructure with fundamentals in Bitcoin technology, based in the appropriate fact that to run, Internet use resources such as numbers and names \cite{hari2016internet}. 

There is no study directly related to this work and there are few blockchain works related to the Internet Infrastructure \cite{angieri2018experiment}. Blockchain still is not a matured technology, there are challenges that need to be considered when designing a platform, to ensure security, reliability and usability. So, there is not related works associated with Internet Infrastructure, because is fact that due to the emergent nature of the topic, the reviewed literature was not published in high-ranking journals with prolonged review cycles \cite{Xu2016}.

The main goal of this paper is to present the \textit{Internet Infrastructure Blockchain} (IIBlockchain), a blockchain architecture to improve the security, knowledge and intelligence in inter-agent communication and collaboration over restrict domains of the \textit{Internet Infrastructure}, developed specifically and therefore independent of any available blockchain platform. The next sections of this paper will be organized as follows. In section 2 we discuss the A2RD model and the needs for inter-agents communication and cooperation. In section 3 we present the architecture of IIBlockchain and the properties inherent to the blocks, their types and the characteristics of the designed chain. In section 4 we discuss the implementation of IIBlockchain showing the main associated properties. In section 5 we present the conclusions and in section 6 we present the proposals for future works.

\section{The A2RD Model}

A2RD is a project that initially proposed the creation of agents with automatic activities replacing human tasks in the environment restricted to the domain of an AS. The use case was the addition and update of objects in IRR server. The application was considered useful mainly because the tasks of the AS administrator did not guarantee the accuracy in its completion nor the permanent need to update the objects making the IRR an unreliable system from the point of view of its contents. A2RD solved this problem.

\begin{figure}[!ht] 
\centering
\includegraphics[scale=0.35]{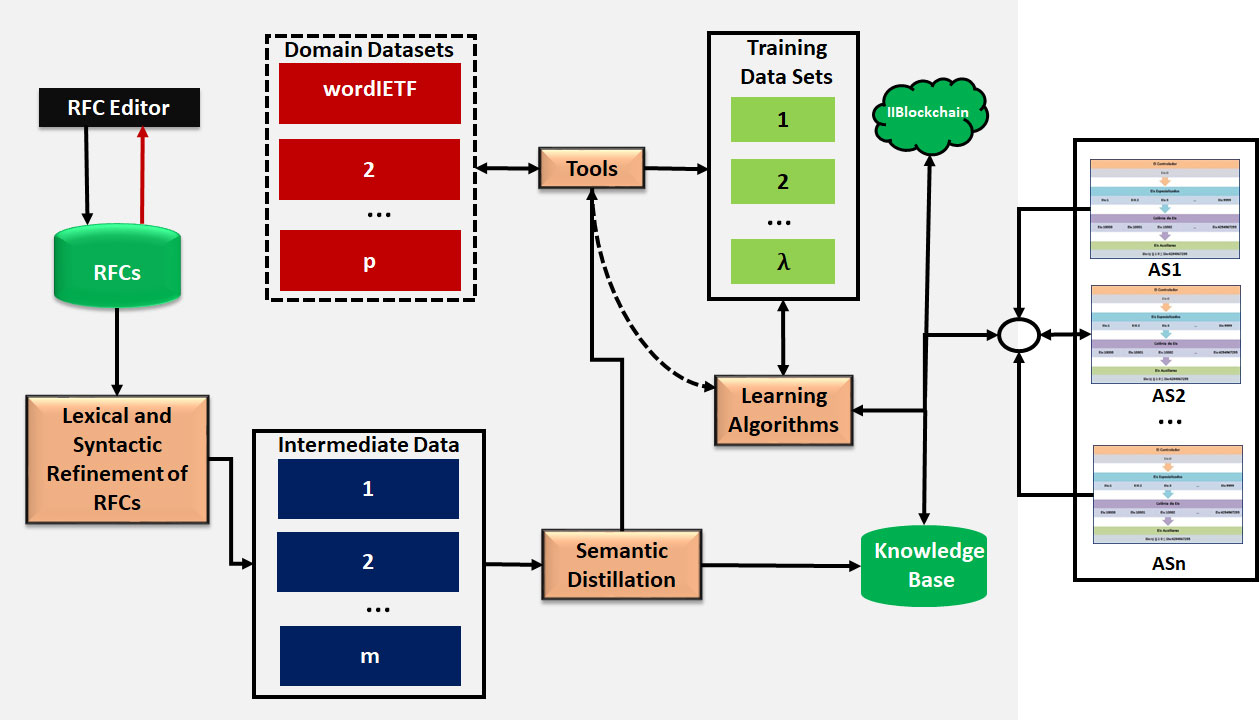}
\caption{A2RD environment}
\label{fig:a2rdEnvironment}
\end{figure}

A new proposal for the A2RD model emerged from this experience  \cite{Braga:2017}. The Figure \ref{fig:a2rdEnvironment} shows this new proposal, in which each implementation of A2RD is represented as an agglomeration of IEs in a four layers model.

A2RD IEs, reach their autonomy and intelligence aided by four components, the \textit{Knowledge Base}, the \textit{Training Data Sets}, and \textit{Domain Datasets}. These first three components are obtained from non-structured databases, in particular, from the \textit{Request for Comments} database, containing documents authored by network operators, engineers and computer scientists, documentary methods, behaviors, research, or innovations applicable to the Internet, all of them, working in groups of the IETF and IRTF, and maintained by RFC-Editor\footnote{https://www.rfc-editor.org/}.

Each AS, of its own free will, may implement its respective A2RD, which is controlled by the IE named \textit{IE Controller}, which receives the identification \textit{x:0}, where \textit{x} is the \textit{AS Number} (ASN).

IEs need to communicate in order to collaborate, learn and cooperate with each other.  This communication needs to be secure, that is, the respective \textit{IE controller} must recognize the origin of each pair in their information exchanges. A mechanism called \textit{Dark Think Security} (DTS) has been proposed to ensure the desired security \cite{Braga:2017b}. Although preliminary implementations have revealed that DTS is indeed secure, it has proved to be complex in implementation. In the search for a simpler alternative included \textit{Pretty Good Privacy} (PGP) \cite{garfinkel1995pgp}. Using PGP, an AS$x$ IE controller that wants to communicate with an AS$y$ IE controller, will use the AS$y$ \textit{public key} to encrypt the message, for $\forall x\quad  and\quad \forall y\quad such\quad that\quad x \neq y\quad and\quad x,\quad y = 1, ..., n,\quad n \leq$ total ASes present in the \textit{Internet Routing Table}\footnote{http://thyme.rand.apnic.net/current/data-summary}. The AS$y$ controller uses AS$x$ \textit{secret key} to decrypt the message. Thus, for this and for other reasons that we will see in the following section,  the recommended solution was a variation of blockchain implementations proposed in the literature, that we named in this paper as \textit{IIBlockchain}, which represents the fourth component, as a cloud, in Figure \ref{fig:a2rdEnvironment}.

%\section{Motivation}
%\jb{ Motivation: Why does this solution is important? What issues does it solve?}

\section{IIBlockchain Model and Implementation}
%\jb{i think it could be very nice to start this section describing main goals of IIBlockchain and how it works at high level. then, you describe it in more technical details}

%\jb{do you have a high level figure to represent the IIBlockchain?}

\begin{figure}[ht]
\centering
\includegraphics[scale=0.45]{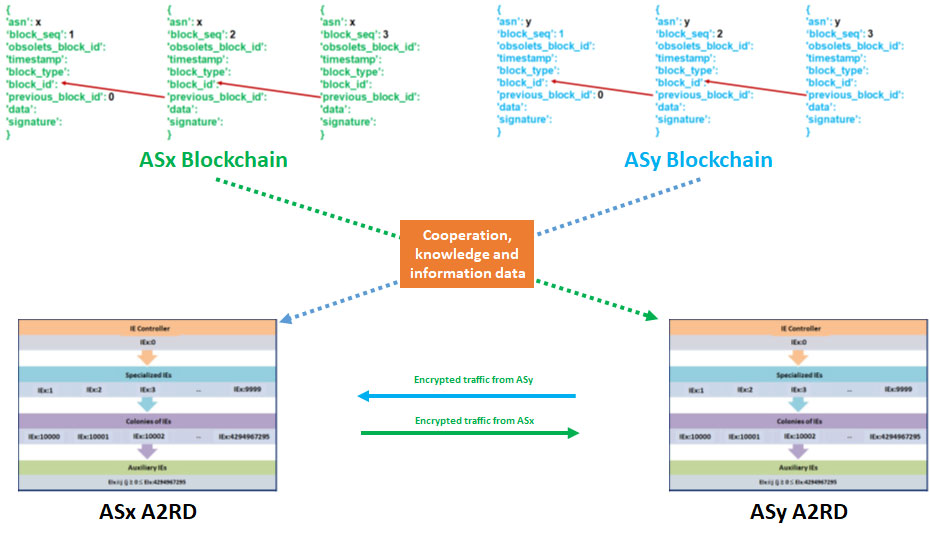}
\caption{IIBlockchain Architecture implemented over ASx and ASy domains}
\label{fig:arquiteturaIIBlockchain}
\end{figure}

The IIBlockchain model can be seen in Figure \ref{fig:arquiteturaIIBlockchain}, which shows the implementation of A2RD in any two ASes (AS$x$ and AS$y$).

This figure shows that the respective A2RD communicate through encrypted messages. Also, the A2RDs independently maintain a blockchain with properties characteristic of IIBlockchain. These chains contain, in their blocks, data inherent to each A2RD and about the environment of the AS in which they are implemented allowing the cooperation through the exchange of knowledge and information that can help in learning and maintaining the autonomy of their respective IEs. Each A2RD locally maintains a copy of IIBlockchain from each of the other ASes. There is no need to implement an A2RD for a chain to be constructed for an ASN. Specialized IEs of an ASx any guarantee that minimal information is included in chains of other ASes.

\subsection{Block Properties}

\begin{figure}[!ht]
\centering
\includegraphics[scale=0.35]{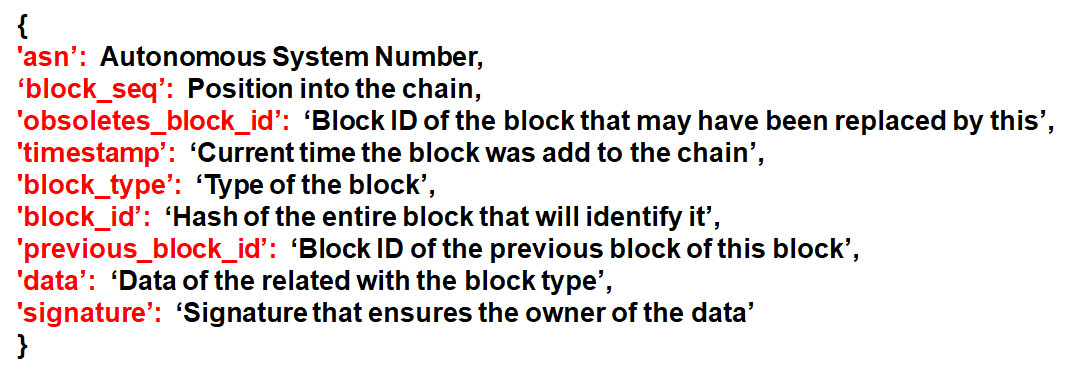}
\caption{Block Structure}
\label{fig:arquiteturaBlockchain}
\end{figure}

A block of any chain type is equivalent to a dictionary structure of the Python language, whose configuration and summary description of the respective keys are shown in Figure \ref{fig:arquiteturaBlockchain}.

The detail description of block keys are in Table \ref{tab:blockKeys}.

\begin{table}[ht]
\centering
\caption{Description of block dictionary keys}
\label{tab:blockKeys}
\begin{tabular}{|l|l|}
\hline
\multicolumn{1}{|c|}{\textbf{Key}} & \multicolumn{1}{c|}{\textbf{Description}} \\ \hline
\textbf{asn} & \makecell[l]{ASnumber: Identifies the owner number of the string.\\For same string, the value of this key is always the same}  \\ \hline
\textbf{block\_seq} & \makecell[l]{Identifies the position of the block within the chain. If\\block $i$ preceding or immediately preceding block $j$ than\\$i < j$ and not necessarily $j = i + 1$. This is due to the fact\\that a block can be removed, from an \textit{ASN} chain, if it\\becomes obsolete. Upon removal, the block is added to\\the \textit{obsolete} chain. The immutability and integrity of\\this \textit{ASN} chain must be restored.}  \\ \hline
\textbf{obsoletes\_block\_id} & \makecell[l]{If the value of this key is not empty, so this references\\the \textbf{block\_id} that will be obsoletes} \\ \hline
\textbf{timestamp} & \makecell[l]{Time moment the block was add in the chain} \\ \hline
\textbf{block\_type} & \makecell[l]{Type of the block: block types are not necessarily\\predefined. IEs can create different types of blocks\\through agreements between them during their normal\\activities. Important blocks are, however, predefined. For\\example, the \textit{Genesis} block, which is necessarily the\\first block of any chain. Blocks that represent IRR\\objects always prefix the usual object name with \textit{irr\_}} \\ \hline
\textbf{block\_id} & \makecell[l]{Hash that will identify the block, obtained on the whole\\block, after it is completely filled} \\ \hline
\textbf{previous\_block\_id} & \makecell[l]{\textbf{block\_id} of the previous block of this block} \\ \hline
\textbf{data} & \makecell[l]{Data of the related with the block type} \\ \hline
\textbf{signature} & \makecell[l]{Signature that ensures the owner of the data} \\ \hline
\end{tabular}
\end{table}

\subsection{Chain Properties}

\begin{figure}[!ht]
\centering
\includegraphics[scale=0.35]{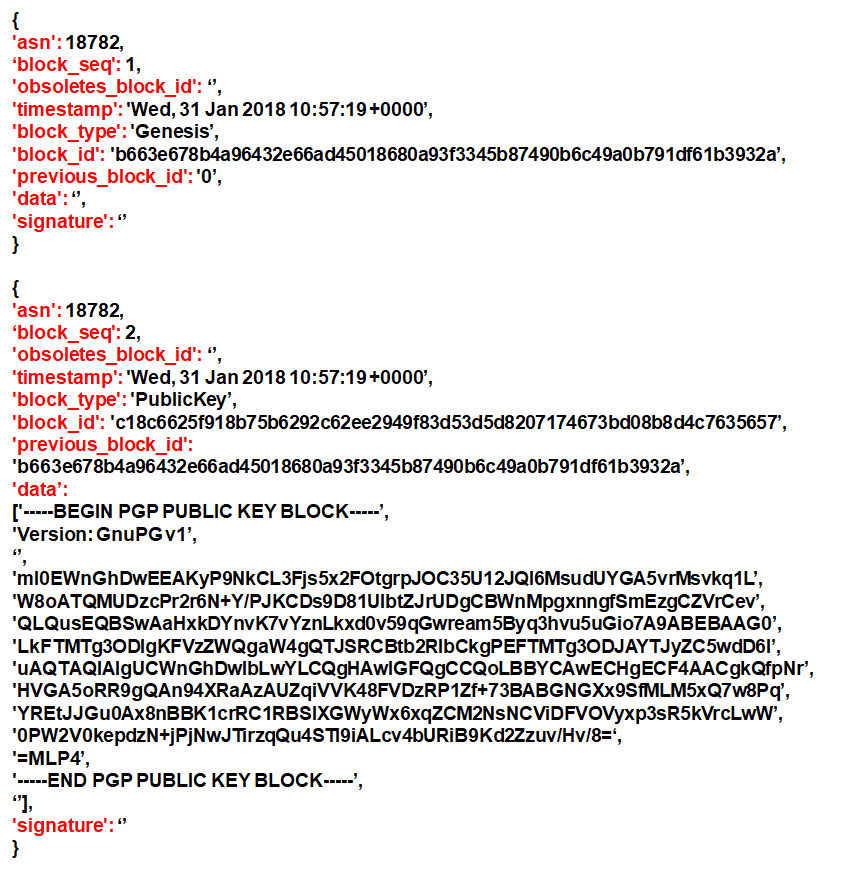}
\caption{Initial chain that, necessarily, has the block type 'Genesis'}
\label{fig:chain-2-blocks}
\end{figure}

Any chain only exists if it has a 'Genesis' block type as its first block ('block\_seq' = 1). Suppose that AS$x$ wants to add in its chain, a block that will contain its \textit{PGP public key} with which any ASN can encrypt messages that only AS$x$ will understand. At this point, the AS$x$ chain is empty. Suppose $x = 18782$. So, using the IIBlockchain Python class available at GitHub\footnote{https://github.com/juliaobraga/a2rd/} if we add the block of type \textit{PublicKey} we will have a two block chain as can be seen in Figure \ref{fig:chain-2-blocks}. It is important to note that block numbers (\textit{block\_seq}) are sequential ($1$ and $2$, respectively).

\begin{figure}[!ht]
\centering
\includegraphics[scale=0.35]{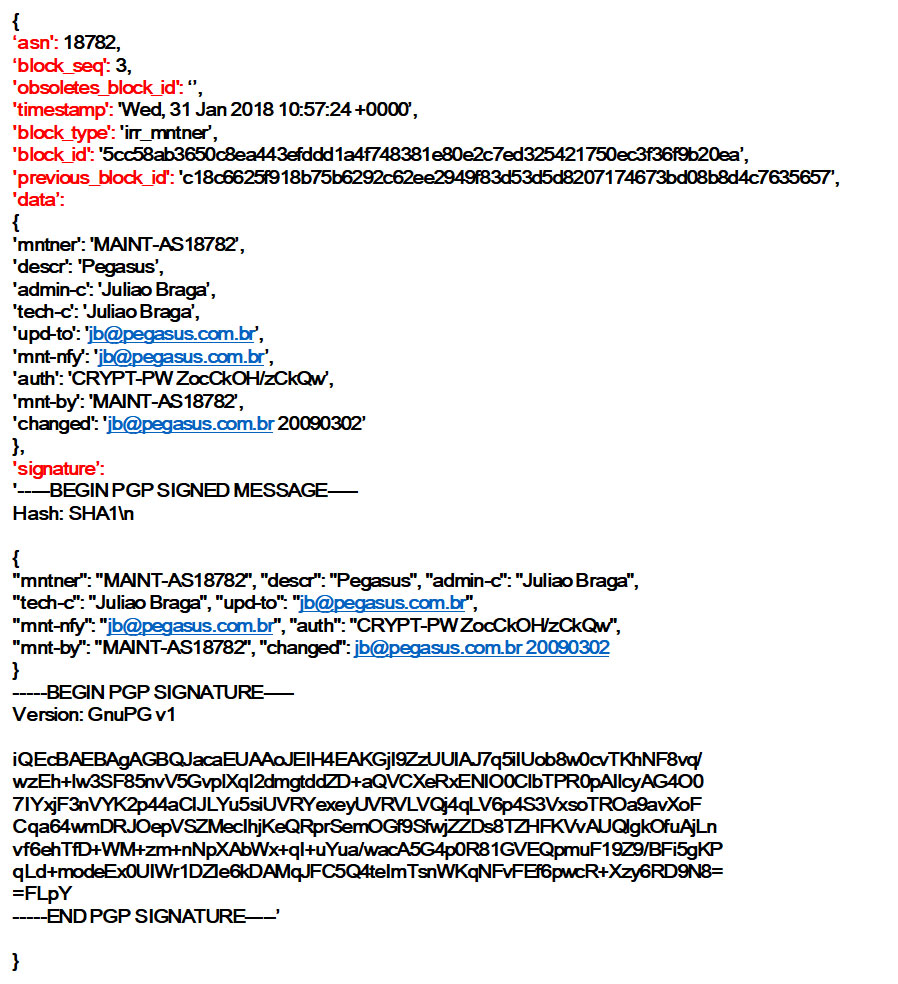}
\caption{Block 3: Adding an IRR object}
\label{fig:block-3}
\end{figure}

Continuing and add a third block, now an \textit{mntner} IRR object type (\textit{irr\_mntner}). The data is transformed into a string to be signed by the PGP, ensuring data properties to AS18782. Once this is done, the block is added to the chain as the third block. The block added can be seen in Figure \ref{fig:block-3}.

\begin{figure}[!ht]
\centering
\includegraphics[scale=0.35]{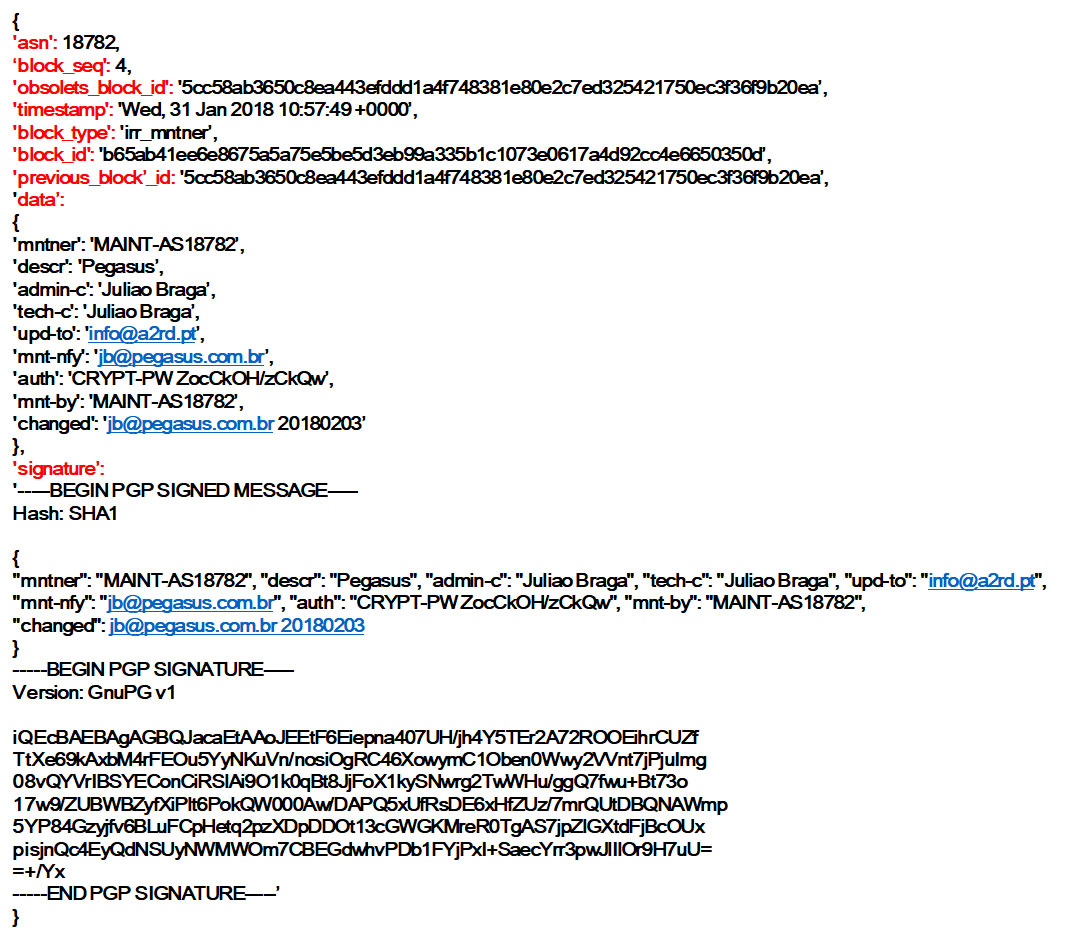}
\caption{Block 4: Makes block 3 obsolete}
\label{fig:block-4}
\end{figure}

To complete these example that illustrate some properties of the chain, let's assume a change in the object \textit{irr\_mntner}. A new data is signed via PGP, and included in the chain, not without first identifying in the \textit{obsoletes \_block\_id}, the block that it is rendering obsolete. The new block is added as 4th block in the AS18782 IIBlockchain and your configuration is shown in Figure \ref{fig:block-4}.

\subsection{Chain Transfer}

The chains are compressed and named as \textit{ASxVaaaammddhhmmss.zip}. A specialized IE will take care of this activity and follow up by compacting the chain, sending it to GitHub\footnote{https://github.com/juliaobraga/a2rd} and update the respective version in \textit{wordIETF}. All chains are public, but the \textit{secret keys} are not.

\section{IIBlockchain Implementation}
%\jb{Discuss about the implementation cost}

In this section we make considerations on important topics that deserved special attention during implementation.

\subsection{Space Analysis}

Table \ref{tab:storage} displays some data about storage values, considering the chain created for the example in this paper.

\begin{table}[!ht]
\centering
\caption{Storage Costs Parameters}
\label{tab:storage}
\begin{tabular}{|l|l|r|}
\hline
\multicolumn{1}{|c|}{\textbf{\#}} & 
\multicolumn{1}{|c|}{\textbf{Discrimination}} & \multicolumn{1}{c|}{\textbf{Value}} \\ \hline
1 & Block 1  &  1,300 \\ \hline
2 & Block 2  &  1,365 \\ \hline
3 & Block 3  & 3,451  \\ \hline
4 & Block 4  & 3,571 \\ \hline
5 & Total & 9,687 \\ \hline \hline
6 & ASes in routing table (12 Fev 2018) & 59,789 \\ \hline \hline
7 & IRR objects number (ARIN) & 10 \\ \hline \hline
8 & Number of protocols in TCP/IP & 51 \\ \hline \hline
\end{tabular}
\end{table}

We used the \textit{sys.getsizeof} function to determine the amount of bytes occupied by the Python dictionary structure, chosen to represent IIBlockchain. The result is not very good and so we evaluated two alternatives versions. The preferred version was that of larger result values\footnote{https://goshippo.com/blog/measure-real-size-any-python-object/} (lines 1-4 on the table). Suppose each block of the string to be constructed occupies twice as many bytes as the largest block in our example (line 4). So our block occupies \textbf{7,124 bytes}. \textit{American Registry for Internet Numbers}\footnote{https://www.arin.net/resources/routing/templates.html} (ARIN) identifies ten objects to populate its IRR (line 7). Thus, only with IRR objects, the IIBlockchain of an AS spends $7,124 \times 10 = \textbf{71,124 bytes} \sim \textbf{70 Kbytes}$. So, the total bytes to represent the IRR objects for all ASn are: $59,789 \times 70$ Kbytes $= 4,285,675,520 \sim \textbf{4 Gbytes}$. Let us now assume that for each TCP/IP protocol\footnote{http://www.comptechdoc.org/independent/networking/protocol/protnet.html} (line 8) we will need 20 blocks with the largest known double size  (knowledge information, for example): $20 \times 7,124 \quad bytes = \textbf{139 Kbytes}$, value that corresponds to $\textbf{0.003\%}$ of the space spent by IRR objects. Certainly there are other types of blocks that IEs will produce. But the largest number of them are obsolete blocks. Very difficult to measure the space to be occupied by obsolete blocks. Only an inaccurate estimate would be possible. One estimate is that 25\% of the blocks will be obsolete. So the total estimated storage space for the IIBlockchain is \textbf{5 Gbytes}. Any operation on IIBlockchain do not require additional space. Therefore the space complexity is \textbf{O(1)} $\sim$ \textbf{O(n)} \cite{costa2015python}.

\subsection{Time Complexity}

The heaviest algorithm we have in operations on IIBlockchain is to search linearly over an array or eventually over a linked list. Then, in the worst case, the complexity of time is O(n) \cite{costa2015python}.

\subsection{Security}

IIBlockchain is public. The security that matters to IIBlockchain will only be verified when a non obsolete block needs to be used. In two stages this is necessary: (a) the integrity of the block and (b) the reliability of the information contained in the block. Stage (a) consists of checking the validity of the hash that identifies the \textit{block\_id}. Stage (b) is the verification that the signature guarantees ownership of the information by the respective AS. If any of the above stages fails, an alert is sent to all implementations of A2RD. Immediately look for the block in the previous version and use it. The existence of the block in the previous version can be verified by the parameter \textit{timestamp} and the name of the version. Meanwhile, specialized IEs will analyze the chain, in order to identify the cause of the breach of trust in the block.

\section{Conclusions}

The authors consider that the objective of allowing a mechanism of relationship between IEs of the various A2RD implementations was achieved. Also, Blockchain is effective in ensuring co-operation and distribution of knowledge that can be shared among IEs in the various domains of ASes. It is a simple, easy-to-understand, and implementation-oriented design with no additional effort required in any programming language. The IIBlockchain has both public and private characteristics and has no inherent concerns or additional difficulties, for this reason. Also, it is worth remembering that IIBlockchain is oriented to the application of Blockchain by agents and not by humans, which certainly decreases complexity.

\section{Future Works}

At this point, it is not possible to determine how the presence of obsolete blocks will influence the operations on an IIBlockchain of some ASN. Implementations in programming languages like Python and others one, does not seem to be a big problem, because dictionaries are indexed and obsolete blocks can be ignored. However, it is necessary to evaluate the possibility of creating a new type of chain: the chain of obsolete blocks, that is to say, the chain consisting of blocks that become obsolete in each ASN chain.

At some point, one A2RD IE may checking the state of the chain and remove obsolete blocks, passing it to the obsolete chain considering:

\begin{itemize}
\item The chain from which the block was removed will be reconstituted to maintain the immutability and integrity. This is achieved by having the next block point to the previous block removed, and a new hash is calculated to identify the next block and successively to the blocks thereof until the end of the chain.
\item The block removed will be inserted in the \textit{obsolete chain} pointed to the last block of this chain. The block's block number ('block\_seq') should be concatenated by a hyphen and another sequence number to the number of the last block of the \textit{obsolete chain}. After this a new hash will be determined to identify this block and the block can be inserted in the \textit{obsolete chain} 
\end{itemize}

Complementary, the IIBlockchain design is simple enough for applications in several other networking areas or not. New versions of the implementation will seek to establish independence from block structure and coding.

\section{Thanks}

From Juliao Braga and Jessica Ribas: Supported by CAPES – Brazilian Federal Agency for Support and Evaluation of Graduate Education within the Brazil’s Ministry of Education.

\bibliographystyle{sbc}
\bibliography{sbc-template}

\end{document}